%% file: main.tex
\documentclass[sigconf]{acmart}

\AtBeginDocument{%
  }

\setcopyright{acmlicensed}
\copyrightyear{2024}
\acmYear{2024}
\acmDOI{XXXXXXX.XXXXXXX}

\setcopyright{acmlicensed}
\copyrightyear{2024}
\acmYear{2024}
\acmDOI{XXXXXXX.XXXXXXX}
\acmConference[KDD WorkShop]{Make sure to enter the correct conference title from your rights confirmation email}{August 25-26, 2024}{Barcelona, Spain}
\acmISBN{978-1-4503-XXXX-X/18/06}

\usepackage{algorithm}  
\usepackage{algorithmicx}  
\usepackage{algpseudocode} 
\usepackage{diagbox}
\usepackage{booktabs}
\usepackage{color}
\usepackage{graphicx}
\usepackage{subcaption}
\usepackage{multirow}
\usepackage{amsmath}
\usepackage{setspace}
\usepackage{enumitem}

\usepackage{amssymb}
\usepackage{amsmath}

\usepackage{xcolor}

\begin{document}

\title{Addressing Graph  Anomaly Detection via Causal Edge Separation and Spectrum}

\author{Zengyi Wo}
\email{wozengyi1999@tju.edu.cn}
\orcid{0009-0007-6939-0042}
\affiliation{%
  \institution{Tianjin University}
  \city{Tianjin}
  \country{China}
}

\author{Wenjun Wang}
\affiliation{%
  \institution{ Tianjin University}
  \city{Tianjin}
  \country{China}
}

\author{Minglai Shao}
\authornote{Corresponding author}
\email{shaoml@tju.edu.cn }
\affiliation{%
  \institution{Tianjin University}
  \city{Tianjin}
  \country{China}
}

\author{Chang Liu}
\affiliation{%
  \institution{Tianjin University}
  \city{Tianjin}
  \country{China}
}

\author{Yumeng Wang}
\affiliation{%
  \institution{Tianjin University}
  \city{Tianjin}
  \country{China}
}

\author{Yueheng Sun}
\authornote{Corresponding author}
\email{yhs@tju.edu.cn}
\affiliation{%
  \institution{Tianjin University}
  \city{Tianjin}
  \country{China}
}

\begin{abstract}
In the real world, anomalous entities often add more legitimate connections while hiding direct links with other anomalous entities, leading to heterophilic structures in anomalous networks that most GNN-based techniques fail to address. Several works have been
proposed to tackle this issue in the spatial domain. However, these methods overlook the complex relationships between node structure encoding, node features, and their contextual environment and
rely on principled guidance, research on solving spectral domain heterophilic problems remains limited. This study analyzes the
spectral distribution of nodes with different heterophilic degrees and discovers that the heterophily of anomalous nodes causes the
spectral energy to shift from low to high frequencies. To address the above challenges, we propose a spectral neural network CES2-
GAD based on causal edge separation for anomaly detection on heterophilic graphs. Firstly, CES2-GAD will separate the original graph into homophilic and heterophilic edges using causal interventions. Subsequently, various hybrid-spectrum filters are used to capture signals from the segmented graphs. Finally, representations from multiple signals are concatenated and input into a classifier to
predict anomalies. Extensive experiments with real-world datasets have proven the effectiveness of the method we proposed.
\end{abstract}

\begin{CCSXML}
<ccs2012>
   <concept>
       <concept_id>10010147.10010257.10010293.10010294</concept_id>
       <concept_desc>Computing methodologies~Neural networks</concept_desc>
       <concept_significance>500</concept_significance>
   </concept>
</ccs2012>
\end{CCSXML}

\ccsdesc[500]{Computing methodologies~Neural networks}

\keywords{ Graph Anomaly Detection; Causal Analysis; Heterophily; Spectrum}

\maketitle

\input{context/introduction}
\input{context/relatedwork}
\input{context/preliminary}
\input{context/method}
\input{context/experiments}
\input{context/conclusion}

\begin{acks}
This work is supported by the National Natural Science Foundation of China (No.62272338) and the National Key R\&D Plan of China (Grant No.2022YFC2602305).
\end{acks}

\bibliographystyle{ACM-Reference-Format}
\bibliography{sample-base}

\end{document}

%% file: context/introduction.tex
\section{Introduction}
\begin{figure}[h!]
    \centering
    \includegraphics[width=0.91\linewidth]{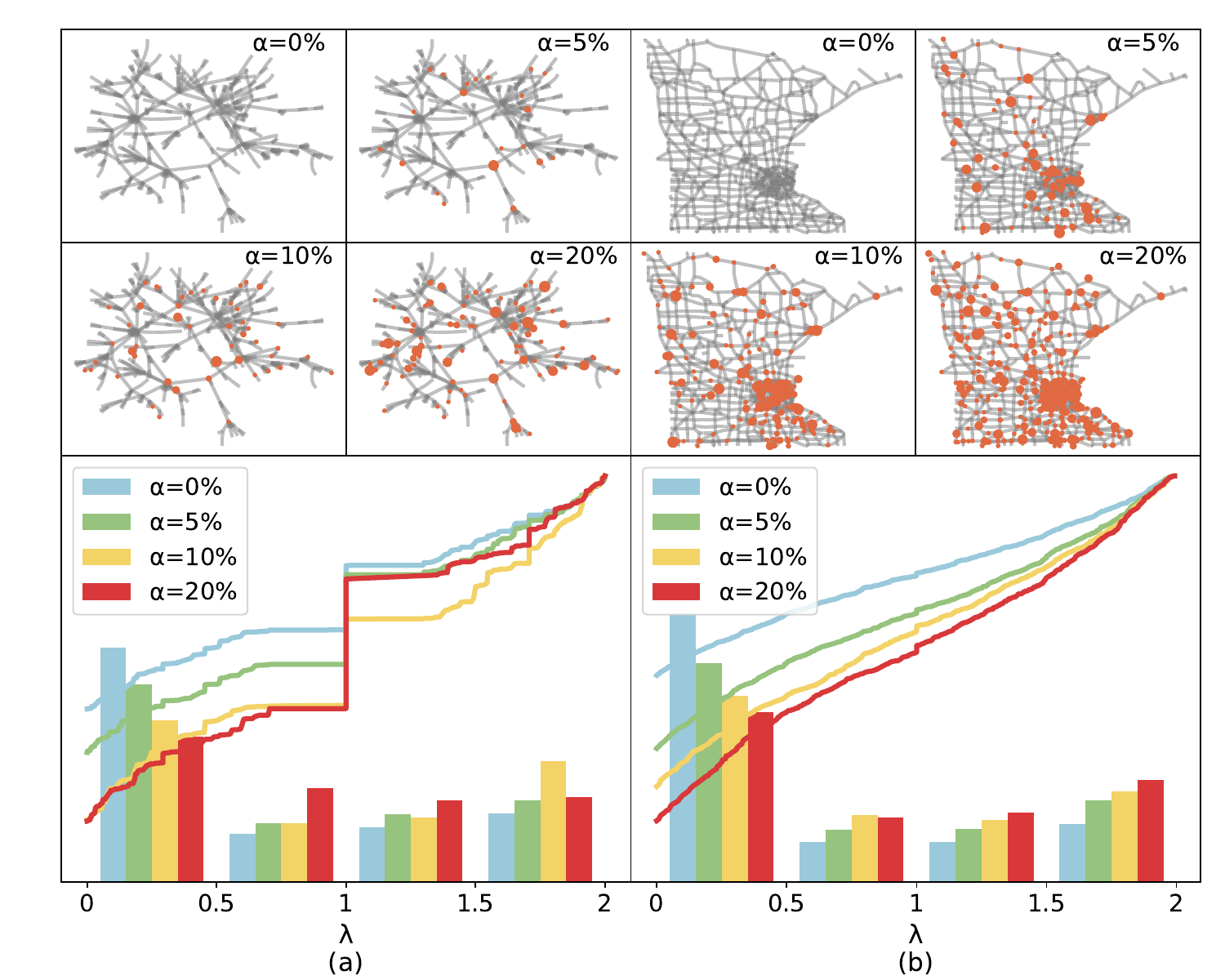}
    \caption{Analyzing the effects of anomalies of different degrees of severity on spatial domain (top) and spectral domain (bottom) graphs, specifically focusing on the Barabási-Albert graph (a) and the Minnesota road graph (b) at various anomaly ratios ($\alpha$ = 0\%, 5\%, 10\%, 20\%).}
    \label{intro}
    \vspace{-0.6cm}
\end{figure}
In the current era of digital technology, the volume of information on the internet is experiencing exponential growth. Concurrently, anomalous activities are increasing in various domains such as financial networks\cite{zhang2022efraudcom}, social networks\cite{deng2022markov}, and academic networks\cite{cho2021masked}, becoming increasingly rampant and causing significant losses across numerous industries. Therefore, accurately detecting potential anomalous activities has become urgent and critically significant. This issue has recently garnered widespread attention from researchers.

Graph Neural Networks (GNNs) have seen widespread application in various fields in recent years, owing to their capacity to effectively model graph-structured data. There is a growing trend in research to integrate GNNs into Graph Anomaly Detection (GAD) systems to capitalize on the intricate topological information present in anomalous graphs. However, anomalous graphs pose a unique challenge due to a combination of homophily and heterophily. Anomalous entities often interact with legitimate users to mask their direct links with other anomalous entities, creating a higher proportion of heterophilic edges in the anomalous graph. This complexity complicates anomaly detection efforts as traditional GNN algorithms operate under the \textit{homophily assumption}, which smooths out node representations during message aggregation and propagation. Existing anomaly detection\cite{dou2020enhancing, liu2020alleviating} approaches based on GNNs and the homophily assumption may introduce inaccuracies in learning representations, ultimately impacting the effectiveness of anomaly detection methods.

Recently, numerous studies have been exploring ways to address the aforementioned challenges\cite{wo2024graph}. \textit{The spatial domain.} These methods aim to alleviate the noise introduced by heterophilic neighbors during the message aggregation process. They seek to address the heterophily problem in anomalous graphs through strategies such as aggregating similar neighbors\cite{liu2020alleviating}, resampling neighbors\cite{dou2020enhancing}, and reweighting edges\cite{cui2020deterrent}. \textit{The spectral domain.} This approach is typically based on spectral GNNs, designing appropriate filters\cite{tang2022rethinking} to distinguish between high-frequency and low-frequency signals. This helps in more effectively handling heterophilic information. These methods overlook the complex relationships between node structure encoding, node features, and their contextual environment. Without guiding principles, this could lead to underperformance and may not be suitable for anomaly detection.

To address the above challenges, we propose a spectral neural network CES2-GAD based on causal edge separation for anomaly detection on heterophilic graphs. Firstly, CES2-GAD will separate the original graph into homophilic and heterophilic edges using causal interventions. Subsequently, various hybrid-spectrum filters are used to capture signals from the segmented graphs. Finally, representations from multiple signals are concatenated and input into a classifier to predict anomalies. Summary of the contributions made in our work:
\begin{itemize}
\item We analyzed the spectral energy distribution of graphs constructed by homophilic and heterophilic edges. We observed that heterophily leads to a shift of spectral energy from low to high frequencies, with high-frequency signals obtaining a more genuine representation after separation. 
\item We propose a spectral neural network CES2-GAD based on causal edge separation for anomaly detection on heterophilic graphs.
\item Extensive experiments on real-world datasets were conducted, and the results indicate that the performance of CES2-GAD surpasses state-of-the-art methods.
\end{itemize}

%% file: context/relatedwork.tex
\section{Related Work}
\subsection{Heterophilic Graph Neural Networks}
Heterophily has emerged as a significant concern for GNNs, and this issue was initially highlighted by \cite{pei2020geom}. Separating ego- and neighbour-embeddings \cite{zhu2020beyond} proves to be an effective technique for learning on heterophilic graphs. Given that nodes sharing the same class are distantly placed within heterophilic graphs, several approaches aim to extend the local neighbors to non-local ones by integrating multiple layers \cite{abu2019mixhop}, and identifying potential neighbors through attention mechanisms \cite{liu2021non} or similarity measures \cite{jin2021node}. Spectral-based methods \cite{luan2021heterophily} overcome this challenge by introducing additional graph filters and mixture strategy, which aims to adaptively integrate information by emphasizing certain frequencies. 

\subsection{GNN-based Anomaly Detection}
To make GNNs, which are based on the \textit{homophily assumption}, more effective for GAD, many researchers have conducted extensive explorations.

Unlike directly aggregating neighboring embeddings, Graph-Consis\cite{liu2020alleviating} designs three modules to address three inconsistency issues in GNNs for GAD simultaneously. 
CARE-GNN\cite{dou2020enhancing} incorporates a label-aware similarity measure for identifying information-rich neighbor nodes and a similarity-aware neighbor selector.
PC-GNN\cite{liu2021pick} incorporates a label-balanced sampler for constructing subgraphs and a neighborhood sampler for selecting candidate neighbor nodes.
GAGA\cite{wang2023label} proposes group aggregation, explicitly augmenting neighbor features with their labels and considering unlabeled neighbors as a distinct class.
IHGAT\cite{liu2021intention} simultaneously captures sequence-like intentions and encodes the relationships between transactions.

%% file: context/preliminary.tex
\section{PRELIMINARY}
\subsection{Background}
\subsubsection{Multi-relation Graph}
Let $\mathcal{G} = \{ \mathcal{V}, \mathrm{X}, \left \{ \mathcal{E}_r^+, \mathcal{E}_r^- \right \}|_{r=1}^R, \mathrm{Y}\}$ denote a multi-relation graph , comprised of a set of $N$ nodes $\mathcal{V} = \left \{ v_1, v_2, \dots, v_N \right \} $, a homophilic edges set $\mathcal{E}_r^+$ and a heterophilic edges set $\mathcal{E}_r^-$. Here, $R$ is the total number of relations, $\mathrm{X} \in \mathbb{R}^{N\times D}$ denotes the node attribute matrix, and $\mathrm{Y} \in \mathbb{R}^{N\times 1}$ denotes the set of labels where $y = 1$ denotes anomalous nodes and $y = 0$ denotes normal nodes.

\subsubsection{Graph-based Anomaly Detection}
The exploration of graph-based anomaly detection pertains to a semi-supervised binary classification problem. Given a multi-relation graph $\mathcal{G}$, the goal is to train a classifier $f_\theta$ based on information from the set of nodes $\mathcal{V}$, the set of edges $\left \{ \mathcal{E}_r^+, \mathcal{E}_r^- \right \}$, and partial labels data  $\mathrm{Y}_\text{train}$ to predict the label of each node, specifically to detect whether a node is anomalous.

\subsubsection{Graph Heterophily} 
For a graph $\mathcal{G}$, an edge is a heterophilic edge if it connects two nodes with different labels (here referring to anomalous node and normal node). An anomalous graph contains both homophilic and heterophilic edges simultaneously. The heterophily of a node can be expressed by the following formula: 
\begin{equation}
heter_{v_i} = \frac{|\{v_j: v_j \in \mathcal{N}\left ( i \right ) |y_i \ne y_j\}|}{|\mathcal{N}\left ( i \right )|}, 
\end{equation}
\noindent where, $\mathcal{N}(i) = \{ v_j \mid (v_i, v_j) \in \mathcal{E} \}$ denotes the neighbor set of node $v_i$. The heterophily of the whole graph can be defined as:
\begin{equation}
heter_{\mathcal{G}}=\frac{|\{e_{ij}:y_i\neq y_j\}|}{| \mathcal{E}|},
\end{equation}
\noindent where, $e_{ij} \in \mathcal{E}$ denotes the edge between node $v_i$ and node $v_j$.

\subsubsection{Graph Spectrum} 
The graph laplacian matrix\cite{kipf2016semi} can be defined in regular form $\boldsymbol{L}=D-A$ or normalized form $\boldsymbol{L}=I-D^{-1/2}AD^{-1/2}$, where $A$ denotes the adjacency matrix of $\mathcal{G}$, $D$ is the degree matrix and $I$ is the identity matrix. \( \boldsymbol{L} \) is a symmetric matrix with eigenvalues, i.e., \( 0 = \lambda_1 \leq \cdots \leq \lambda_N \) , corresponding eigenvectors \( U = (u_1, u_2, \cdots, u_N) \) and $\hat{x}$ as the graph Fourier transform of $x $. By providing an arbitrary threshold \( \lambda_k \), we can split the eigenvalues into low-frequency \( \{\lambda_1, \lambda_2, \cdots, \lambda_k \} \) and high-frequency \( \{\lambda_{k+1}, \lambda_{k+2}, \cdots, \lambda_N \} \).

To provide a theoretical analysis of the distribution of anomalous graphs, we provide the following definitions:
\begin{itemize}[leftmargin=*]
\item \textit{Spectral energy distribution }at \(\lambda_k\) is defined as:
${\hat{x}_k^2}/{\sum_{i=1}^N \hat{x}_i^2}.$

\item \textit{Spectral energy ratio} at \(\lambda_k\) is defined as the accumulated spectral energy distribution of the first \( k \) eigenvalues, i.e.,$\eta_k(x, L)={\sum_{i=1}^k\hat{x}_i^2}/{\sum_{i=1}^N\hat{x}_i^2}.$
\item \textit{High-frequency Area} can be defined as:
\begin{equation}S_{\text{high}}=\frac{\sum_{k=1}^N\lambda_k\hat{x}_k^2}{\sum_{k=1}^N\hat{x}_k^2}=\frac{x^TLx}{x^Tx}.\end{equation}
\end{itemize}

\subsection{Heterophily and Spectrum with GAD}
\label{3.2}

\subsubsection{Observation}
In the top part of Figure \ref{intro}, we use orange to represent abnormal nodes in the spatial domain. The more orange nodes there are, the greater the degree of anomaly. In the bottom part of the figure, we display the energy distribution of $x$ in the spectral domain across various anomaly ratios. It can be observed that as the heterophilic of anomalous nodes increases, the spectral energy of the low-frequency part decreases significantly, while the spectral energy of the high-frequency part increases significantly. Considering the rarity of anomalous nodes and the relative stability of heterophilic in the entire graph, this is rather remarkable.

\subsubsection{Analysis}
According to spectral graph theory, given the graph Laplacian matrix $\boldsymbol{L}$ and class labels as the ground truth signal $Y$, the high-frequency area $S_{\text{high}}$ of the graph monotonically increases with the heterophily degree of anomalous nodes.
\begin{equation}\begin{aligned}
S_{\text {high}}=\frac{y^TLy}{y^Ty}& =\frac12\sum_{v_i\in V}\sum_{v_j\in V} A_{i,j}\left(y_i-y_j\right)^2/y^Ty,
\end{aligned}\end{equation}
the heterophily of anomalous nodes indicates that anomaly tends to conceal the \textit{anomaly-anomaly }edge and build more \textit{benign-anomaly} edges. Therefore, the high-frequency area $S_{\text{high}}$ will increase with the increase of heterophily.

Given the homophilic subgraph $\mathcal{G}_r^+$ with its Laplacian matrix $\boldsymbol{L}_r^+$, and the heterophilic subgraph $\mathcal{G}_r^-$ with its Laplacian matrix $\boldsymbol{L}_r^-$. The high-frequency area \( S_{\text{high}} \) of the heterophilic subgraph $\mathcal{G}_r^-$ is larger than the homophilic subgraph $\mathcal{G}_r^+$, i.e., 
$S_{\text{high}}(\mathcal{G}_r^-) > S_{\text{high}}(\mathcal{G}_r^+).$

%% file: context/method.tex
\section{METHODOLOGY}
\noindent\textbf{Overview}
As discussed in section \ref{3.2}, we emphasize the importance of high-frequency signals in GAD by analyzing graph distributions with heterophily.
To efficiently extract these signals, We propose a method to separate edges in the original graph and use hybrid filters to learn node representations for anomaly detection, named CES2-GAD. 

Our model consists of three main modules: (1) Causal Edge Separation, (2) Hybrid Graph Spectral Filters, and (3) Anomaly Detection. In the \textit{causal edge separation} module, causal intervention methods differentiate between homophilic and heterophilic edges using supervised signals from labeled nodes. This process helps to separate the original graph into homophilic and heterophilic edges. The \textit{hybrid graph spectral filters} module applies specific filters to these separated graphs to capture diverse signals and combine representations from various relationships. Finally, an MLP classifier is employed to identify abnormal nodes based on the obtained node representations. The following sections will present a detailed overview of each module.

\subsection{Causal Edge Separation}
To successfully separate homophilic and heterophilic edges from the original graph, we propose a causal intervention method. By studying the graph generation process, we aim to estimate the effect of heterophilic treatment on the edge probabilities between two nodes, so that we can intervene in the graph separation by manipulating the heterophilic treatment. 

\subsubsection{Causal Treatment}
We define \( T_{ij} \) as the homophily of the node pair \( (v_i, v_j) \), indicating whether these two nodes exhibit homophily. When \( T_{ij} = 1 \), it means there is an edge between similar node pairs \( (v_i, v_j) \) and no edge between dissimilar node pairs \( (v_i, v_j) \), otherwise $ T_{ij} = 0 $.

\subsubsection{Counterfactual Outcome}
We will now explain the approach used to produce counterfactual results. Since we can only witness the progression of facts and results for each pair of nodes,
therefore, our goal is to estimate counterfactual outcomes that closely mirror the observed results. Essentially, our aim, using a specific set of observed node pairs, is to pinpoint pairs that are most alike in terms of undergoing opposite processes and label them as counterfactual node pairs. For each node pair \( (v_i,v_j) \), we define their counterfactual connection as follows:
\begin{equation}(v_a,v_b)=\arg\max_{v_a,v_b\in\mathcal{V}}\left[s(\tilde{\mathbf{x}}_i,\tilde{\mathbf{x}}_a)+s(\tilde{\mathbf{x}}_j,\tilde{\mathbf{x}}_b)\mid T_{ab}=1-T_{ij}\right],\end{equation}
where $s(\cdot,\cdot)$ is implemented as the Euclidean similarity in the embedded space $\tilde{X}$. Empirically, the embedding $\tilde{X}$ is obtained by connecting the original features of the nodes with their structural encodings (obtained through Laplacian structural encoding), with little influence from homophily.

\subsubsection{Edge Classifier}
With guidance from factual and counterfactual outcomes, we now can learn an edge classifier, which aims to generate augmented views by controlling the treatment variable. 
Given the intermediate node representations $Z$ from our hybrid encoder (detailed later) and the treatment indication $T$ or $T^{CF}$, we can separate a homophilic subgraph $\mathcal{G}_r^+$ and a heterophilic subgraph $\mathcal{G}_r^-$as follows:
\begin{equation}\mathcal{G}_r^+=g_\theta(Z,T),\quad\mathcal{G}_r^-=g_\theta(Z,T^{CF}),\end{equation}
where $g_{\theta}(\cdot, \cdot)$ denotes the edge classifier, and we empirically adopt a simple multi-layer perceptron(MLP).

\subsection{Hybrid Graph Spectral Filters}
After separating the edges of the original graph, the homophily graph represents more low-frequency signals,  while the heterophily graph highlights high-frequency signals. Various filters are then employed on the separated graphs to capture signals across different frequency ranges. The primary objective is to capture common information shared among similar nodes from both homophily and heterophily perspectives. 

\subsubsection{Low-Pass Graph Spectral Filter}
To capture the homophilic view that connects similar nodes, we normalize the adjacency matrix as \(\tilde{\mathcal{E}_r^+} = \widehat{D}^{-1/2} \mathcal{E}_r^+ \widehat{D}^{-1/2}\) and utilize low-pass graph filters in SGC to smooth node representations as follows:
\begin{equation}\label{9}
Z_{l}^{\text{homo}} = \tilde{\mathcal{E}_r^+} Z_{l-1}^{\text{homo}} W^{\text{low}}_{l-1}, \quad \text{and } Z_{0}^{\text{homo}} = X,
\end{equation}
where \(l \in \{1, \ldots, L\}\) denotes the layer index. The final-layer representation matrix for the homophilic view is \(Z^{\text{homo}} = f^{\text{low}}_{\phi}(X, \mathcal{E}_r^+)\), where \(f^{\text{low}}_{\phi}(\cdot, \cdot)\) is the low-pass encoder with parameters \(\phi = \{W_l \mid l \in \{0, \ldots, L-1\}\}\).

\subsubsection{High-Pass Graph Spectral Filter}
To enhance the representation of heterophilic connections between dissimilar nodes, we utilize high-pass filters to diversify node representations. This process can be defined as follows:
\begin{equation}\label{10}
Z_{l}^{heter} = (I - \alpha \tilde{\mathcal{E}_r^-})\ Z_{l-1}^{heter}\ W^{high}_{l-1}, \quad \text{and } Z_{0}^{heter} = X, 
\end{equation}
here, $\tilde{\mathcal{E}_r^-}^{CF}$ represents the normalized ${\mathcal{E}_r^-}$, and the parameter $\alpha$ serves as a hyperparameter governing the intensity of filtering. The resulting heterophilic representations generated by the high-pass encoder are denoted as $Z^{heter}=f^{high}_{\psi}(X, \mathcal{E}_r^-)$.
The node embeddings $Z$ used for downstream tasks is a concatenation from both branches, i.e., $\mathbf{z}_i=[\textbf{z}_{i}^{homo}, \textbf{z}_{i}^{heter}]$.
 This is rooted in previous observations that even for heterophilic edges, neighborhoods could exhibit higher label homophily than others. In such scenarios, comparing representations derived from low-pass filters with those from high-pass filters becomes essential for generating high-quality results. Conversely, when dealing with homophilic edges, the use of high-pass filters can help mitigate issues related to over-smoothing.

\subsection{Anomaly Detection}
We employ cross-entropy loss in the context of anomaly detection. For a given training node set denoted as $\mathcal{V}_{\text{train}}$, where the ultimate embedding of a node $v$ is $\mathbf{z}_v$ and its label is $y_v$, the loss function is formulated as:
\begin{equation}\mathcal{L}_{N}=-\sum_{\boldsymbol{v}\in\boldsymbol{V}_\text{train}}\left[y_{\boldsymbol{v}}\log\left(p_{\boldsymbol{v}}\right)+\left(1-y_{\boldsymbol{v}}\right)\log\left(1-p_{\boldsymbol{v}}\right)\right],
\end{equation}
$p_{\boldsymbol{v}}$ is the softmax value of the embedding vector of node $v$.

%% file: context/experiments.tex
\section{Experiments}
\subsection{Experimental Settings}
\subsubsection{Datasets}
We conducted experiments using the anomaly detection dataset in a real-world scenario to assess CES2-GAD's effectiveness, particularly on the Amazon dataset\cite{mcauley2013amateurs}. This dataset comprises product reviews within the musical instruments category. You can find detailed statistics of the datasets in Table \ref{Dataset statistics}.

\begin{table}[]
    \centering
    \caption{The Statistic of Datasets.}
    \begin{tabular}{c|ccc}
        \hline
        \hline
        {Dataset} & \multicolumn{3}{c}{Amazon}\\
        \hline
        {\#Nodes (Anomaly \%)} & \multicolumn{3}{c}{11944 (6.87\%)}\\
       {Edge-Relations} & U-P-U&U-S-U &U-V-U\\
       {\#Num-Relations} &351216& 7132958 &2073474\\
       Class& Positive& Negative & Unlabeled\\
       {\#Num-Class}&821&7818&3305\\
        \hline
        \hline
    \end{tabular}
    \label{Dataset statistics}
    \vspace{-0.3cm}
\end{table}

\subsubsection{Baselines}
We have conducted a comparative analysis of CES2-GAD against various cutting-edge GNN-based approaches to confirm its efficacy in anomaly detection. 
The first group consists of methods that are \textbf{Feature-based Methods}: including SVM and MLP. 
The second group consists of methods that are based on 
 \textbf{Homophily GNN}: including GCN\cite{kipf2016semi}, GraphSAGE\cite{hamilton2017inductive}, GAT\cite{velickovic2017graph}, GIN\cite{xu2018powerful}, and ChebyNet\cite{defferrard2016convolutional} ;
The third group consists of methods that are \textbf{GNN-based Anomaly Detection (GAD Algorithm)}: including GraphCon-sis\cite{liu2020alleviating}, CARE-GNN\cite{dou2020enhancing}, PC-GNN\cite{liu2021pick}, and GAGA\cite{wang2023label};
The last group consists of methods based on \textbf{Heterophily GNN}: including ACM\cite{luan2022revisiting}, H$^2$-FDetector\cite{shi2022h2}, BW-GNN\cite{tang2022rethinking}, and GDN\cite{gao2023alleviating}.

\subsubsection{Evaluation Protocol}
We selected two commonly used metrics, F1-macro and AUC, to assess the performance of all approaches. F1-macro represents the average F1 score across all classes, disregarding the imbalance between normal and abnormal labels, whereas AUC stands for the area under the ROC curve.

\subsubsection{Model Hyperparameters and Implementation Details}
In our experimental setup, we performed all experiments with PyTorch and an Nvidia 3090 GPU to maintain consistency in the computational environment. The learning rate for the baseline method was determined according to the values specified in the original paper. We partitioned the dataset into training, validation, and testing sets with ratios of 0.4/0.2/0.4 for all methods under comparison. Our methodology was created by leveraging the DGL and PYG libraries within the PyTorch framework, while alternative approaches were coded using publicly available source material.

\subsection{Experimental Results}
\begin{figure*}[h!]
    \centering
    \includegraphics[width=0.85\linewidth]{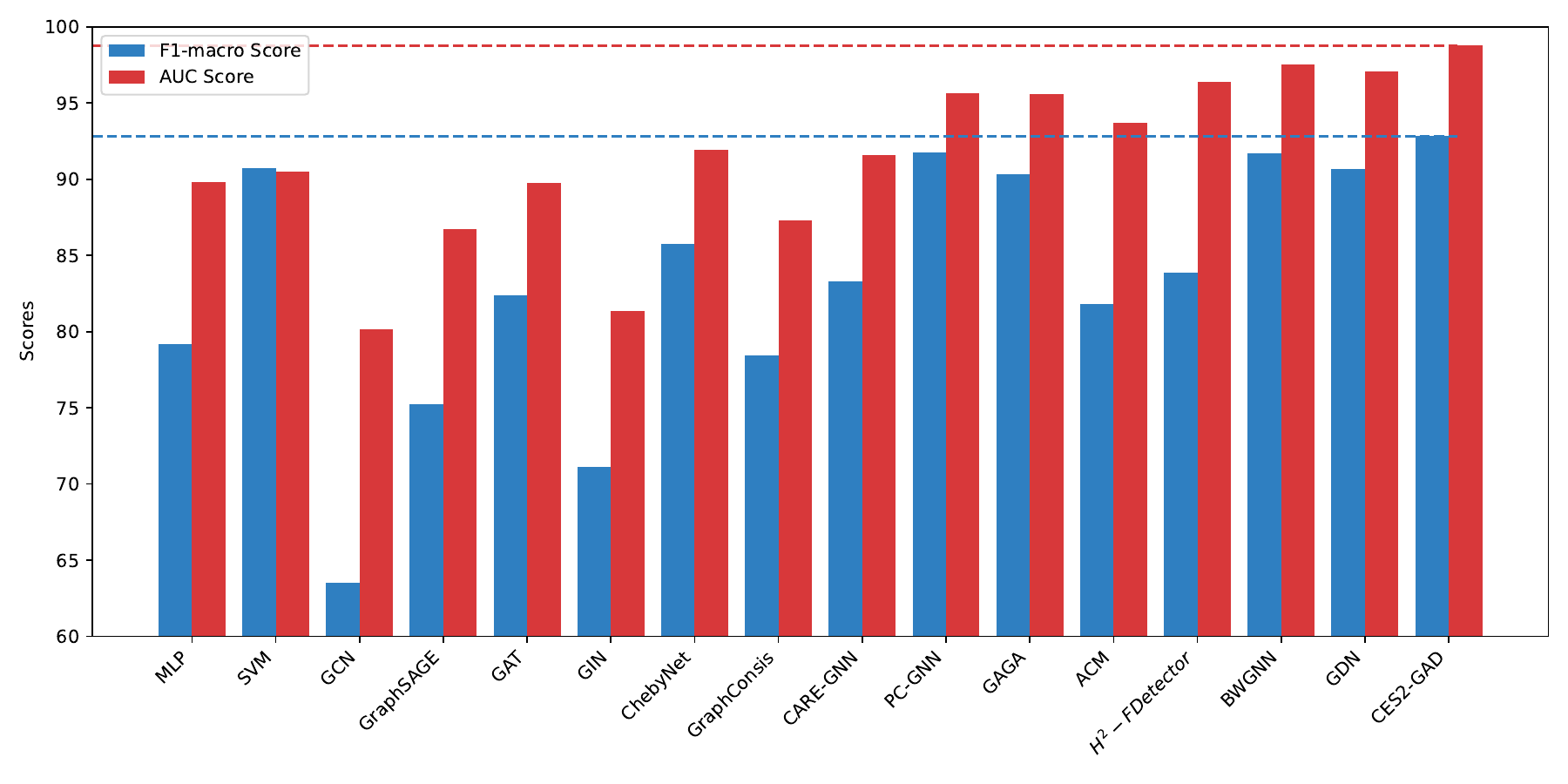}
    \caption{Anomaly detection performance on real-world datasets in terms of F1-macro and AUC value.}
    \label{experiment result}
    \vspace{-0.5cm}
\end{figure*}

We conducted a comprehensive comparison of our method with feature-based methods, conventional Homophily GNN models, state-of-the-art graph-based anomaly detection models, and novel GNNs for Heterophily. The corresponding results are shown in Figure \ref{experiment result}. 
\begin{itemize}[leftmargin=*]
\item \textbf{Feature-Based \& Homophily GNN Methods}: Compared to feature-based methods (SVM and MLP), GNNs following the \textit{Homophily assumption} (GCN, GraphSAGE, GAT, and GIN) showed relatively poor performance, indicating that heterophily problems hinder these general GNNs. Fundamentally, these models can be viewed as \textit{low-pass filters}, focusing only on low-frequency information. 
\item \textbf{GAD Algorithm}: Compared to state-of-the-art graph-based anomaly detection methods, these methods still have significant gaps compared to our proposal. Among them, GraphConsis showed the worst comprehensive performance. It combines node features and the surrounding environment to learn a relationship between attention weight, essentially acting as a low-pass filter. CARE-GNN and PC-GNN showed some improvements in performance because they recognize heterophily edges and the "camouflage" problem, designing corresponding neighbor selection modules. GAGA introduced a group aggregation module to generate distinguishable multi-hop neighborhood information. Despite showing substantial overall improvements, it still exhibited significant gaps compared to our proposed method. 
\item \textbf{Heterophily GNN}: Compared to novel heterophily GNNs, where ACM studies heterophily based on feature similarity, H$^2$-FDetector identifies homophily and heterophily connections and applies different aggregation strategies, BWGNN designs band-pass filters to transmit information in multiple frequency bands, and GDN separates node features into anomaly patterns and neighborhood patterns. The superior performance of CES2-GAD demonstrates the effectiveness of the proposed graph \textit{edge separation} based on causality and \textit{hybrid graph spectral filter} modules.
\end{itemize}

%% file: context/conclusion.tex
\section{conclusion}
This article explores capturing hidden anomalies in heterophilic graphs. We first analyze the relationship between the heterophily of anomalies in the spatial domain and high-frequency signals in the spectral domain. After observing that higher heterophily leads to spectral energy shifting from low to high frequencies, we further propose using causal intervention to separate heterophilic graphs and obtain a more substantial representation, filtered through a hybrid blending filter to capture anomalies in heterophilic graphs. Extensive experiments on real-world anomaly detection datasets demonstrate the effectiveness of CES2-GAD.